\documentclass{article}

\usepackage{iclr2027_conference,times}

\usepackage{amsmath,amsfonts,bm}

\def\eqref#1{equation~\ref{#1}}

\def\1{\bm{1}}

\DeclareMathAlphabet{\mathsfit}{\encodingdefault}{\sfdefault}{m}{sl}
\SetMathAlphabet{\mathsfit}{bold}{\encodingdefault}{\sfdefault}{bx}{n}

\usepackage{amsmath}
\usepackage{amssymb}
\usepackage{booktabs}
\usepackage{graphicx}
\usepackage{multirow}
\usepackage{makecell}
\usepackage{subcaption}
\usepackage{wrapfig}
\usepackage{array}
\usepackage{colortbl}
\usepackage{placeins}
\usepackage{url}
\usepackage[hidelinks]{hyperref}
\usepackage{cleveref}
\crefname{figure}{Figure}{Figures}
\Crefname{figure}{Figure}{Figures}
\crefname{table}{Table}{Tables}
\Crefname{table}{Table}{Tables}
\title{MoETok: Sparse Expert Routing for 1D Visual Tokenizers}

\author{Jiahao Guo\textsuperscript{1,\S}, Bao Tang\textsuperscript{1,\S}, Haoxiang Cao\textsuperscript{3,\S}, Jingfeng Yao\textsuperscript{1},\\
\bfseries Bo Li\textsuperscript{2}, Wenyu Liu\textsuperscript{1}, Kun Gai\textsuperscript{2}, Changqian Yu\textsuperscript{2,\textdaggerdbl}, Xinggang Wang\textsuperscript{1,\textdagger}\\[0.5em]
\normalfont\textsuperscript{1}Huazhong University of Science and Technology\quad
\textsuperscript{2}KlingAI Research\\
\normalfont\textsuperscript{3}South China Normal University}

\newcommand{\methodname}{MoETok}

\definecolor{moeteal}{HTML}{1B998B}
\definecolor{moeblue}{HTML}{2F6B9A}
\definecolor{lightrow}{HTML}{EEF7F5}
\definecolor{roleheader}{HTML}{EEF1F3}
\definecolor{rolebase}{HTML}{F4F5F6}
\definecolor{rolegreen}{HTML}{EAF5EE}
\definecolor{rolegreentext}{HTML}{24724A}
\definecolor{rolered}{HTML}{FBEDEC}
\definecolor{roleredtext}{HTML}{B34B43}
\definecolor{roleblue}{HTML}{1A7FBB}
\definecolor{roleorange}{HTML}{E69F00}

\newcommand{\detailtext}[1]{\textcolor{roleblue}{#1}}
\newcommand{\contenttext}[1]{\textcolor{roleorange}{#1}}

\iclrfinalcopy

\begin{document}

\maketitle
\begingroup
\renewcommand{\thefootnote}{\fnsymbol{footnote}}
\footnotetext[4]{Work done during internship at KlingAI Research.}
\footnotetext[2]{\fontsize{8}{9.5}\selectfont Corresponding author: X. Wang ({\scriptsize\texttt{xgwang@hust.edu.cn}}).\enspace\textsuperscript{\textdaggerdbl}Project leader: C. Yu ({\scriptsize\texttt{y-changqian@outlook.com}}).}
\endgroup
\begingroup
\renewcommand{\thefootnote}{}
\footnotetext{Project page: \url{https://gjhhust.github.io/moetok-project}.}
\endgroup
\lhead{}

\begin{abstract}
Modern visual tokenizers aim to preserve fine-grained visual details while learning high-level semantic representations for downstream generation and understanding. Query-based 1D visual tokenizers encode images into flexible latent sequences, yet how individual learnable queries gather and encode image information remains poorly understood. We reveal that these queries spontaneously develop complementary roles under semantic supervision: a small set of Content queries captures high-level semantic information, while a larger set of Detail queries preserves fine-grained visual details for reconstruction. To adapt token processing to these heterogeneous roles, we introduce \methodname, a 1D visual tokenizer that routes tokens through sparsely activated experts. Jointly optimized with reconstruction and semantic objectives, the router learns role-specialized processing paths adapted to each token's representation. Across model scales and token budgets, \methodname{} consistently improves reconstruction, semantic representation, and autoregressive generation over dense tokenizers under matched supervision and active computation. On ImageNet, it combines strong reconstruction and semantic representation with state-of-the-art class-conditional generation within the 1D tokenizer family. We further validate \methodname{} at scale in a unified autoregressive model jointly trained for understanding and generation with a next-token cross-entropy objective. The resulting model combines competitive visual understanding with state-of-the-art generation among unified discrete models, matching or surpassing diffusion-based models such as BAGEL and Qwen-Image on benchmarks including DPG-Bench (87.61), GenEval2 (40.04 GM), and WISE (0.49).
\end{abstract}

\section{Introduction}
\label{sec:introduction}

Modern visual tokenizers aim to preserve fine-grained visual details while learning high-level semantic representations for downstream generation and understanding. Early tokenizers were primarily trained for image reconstruction~\citep{vqvae,vqgan}, while recent methods introduce semantic supervision to improve the learned latent space and support both generation and understanding~\citep{vavae,maetok,unitok2025,qlip2025,vtp2025}. TiTok~\citep{titok} introduced 1D visual tokenization, using learnable queries to encode an image into a latent sequence. Subsequent methods extend this formulation through diverse forms of semantic supervision and demonstrate improved representation quality and scaling behavior~\citep{gigatok,beyer2025highly,semanticist,dinotok}. These advances are largely evaluated through the performance of the complete token sequence, leaving a fundamental question: \emph{how do individual learnable queries gather and encode image information?}

We investigate this question through encoder attention analysis~\citep{doshi2026biorthogonal} and per-query linear probe evaluation~\citep{beyer2025highly}. Across 1D tokenizers with diverse architectures and training objectives, semantic supervision promotes spontaneous query-role differentiation: a small subset of queries adapts its attention to image content and contributes strongly to classification, while most queries attend to stable spatial regions. Reconstruction and subset interventions further establish their functional significance: the small Content subset retains high-level semantic information, whereas Detail queries preserve fine-grained visual details for faithful reconstruction. Together, these analyses reveal complementary query roles within the 1D bottleneck.

These findings motivate allocating parameter capacity according to the different information roles of visual tokens. Sparse expert routing provides token-dependent access to a larger parameter pool while activating only a subset for each token. We investigate this design in 1D tokenizers under a fixed token budget and comparable active computation.

Specifically, we introduce \textbf{\methodname}, a 1D visual tokenizer with sparse expert routing. \methodname{} retains shared self-attention and replaces each dense encoder feed-forward network (FFN) with a pool of experts. A learned router selects a small subset of experts for each visual token based on its current representation. Trained jointly with reconstruction and semantic objectives, the router learns expert preferences for different token roles without explicit role labels. The resulting role-specialized processing paths allow \methodname{} to adapt how visual information is processed as the 1D bottleneck forms.

We evaluate \methodname{} through controlled comparisons, ImageNet benchmarks, and large-scale multimodal modeling. Across Small, Base, and Large architectures and both 64- and 256-token bottlenecks, \methodname{} consistently improves reconstruction, semantic representation, and autoregressive generation over dense tokenizers. On ImageNet, it combines strong reconstruction and semantic representation with state-of-the-art class-conditional generation within the 1D tokenizer family. We further use \methodname{} in a unified autoregressive model that performs visual understanding and generation through next-token prediction~\citep{wang2024emu3,emu35}. The resulting unified autoregressive model delivers competitive visual understanding and generation performance comparable to or exceeding BAGEL and Qwen-Image across the evaluated benchmarks~\citep{deng2025emerging,wu2025qwen}. These results demonstrate the effectiveness of sparse expert routing for learning visual representations that support both understanding and generation.

Our contributions are summarized as follows:
\begin{itemize}
    \item We systematically study how learnable queries gather and encode image information in 1D visual tokenizers. Cross-tokenizer analysis reveals emergent Content and Detail queries, whose complementary roles in semantic representation and fine-grained reconstruction are validated through linear probe evaluation and controlled interventions.
    \item We introduce \methodname, a 1D visual tokenizer that learns role-specialized sparse processing paths for heterogeneous visual tokens. It consistently improves reconstruction, semantic representation, and autoregressive generation across model scales and token budgets, achieving state-of-the-art class-conditional generation within the 1D tokenizer family on ImageNet.
    \item We further extend \methodname{} to unified autoregressive modeling, achieving competitive visual understanding and state-of-the-art generation among unified autoregressive models across DPG-Bench, WISE, and GenEval2.
\end{itemize}

\section{Related Work}
\label{sec:related}

\paragraph{1D visual tokenizers.}
TiTok represents images as compact sequences of learnable queries, establishing an alternative to spatially arranged latent grids~\citep{titok}. Subsequent work has advanced adaptive token allocation, continuous representations, and hierarchical token organization for efficient reconstruction and generation~\citep{elastictok2025,alit2025,chen2025softvqvae,spectralar2025,restok2026}. FlexTok uses nested dropout to learn ordered token sequences whose prefixes capture progressively finer image content~\citep{flextok2025}. TA-TiTok incorporates CLIP text features into the decoder, introducing textual conditioning during tokenizer training~\citep{tatitok2025}. Semanticist combines continuous tokens with a PCA-like hierarchy and a diffusion decoder to organize complementary semantic and fine-grained visual information~\citep{semanticist}. GigaTok demonstrates the importance of semantic regularization for scaling 1D tokenizers~\citep{gigatok}, while ReTok improves sequence utilization through redundant token padding and hierarchical semantic regularization~\citep{retok2026}. We investigate emergent query roles and introduce adaptive processing paths for them.

\begin{figure}[t]
    \centering
    \includegraphics[width=\linewidth]{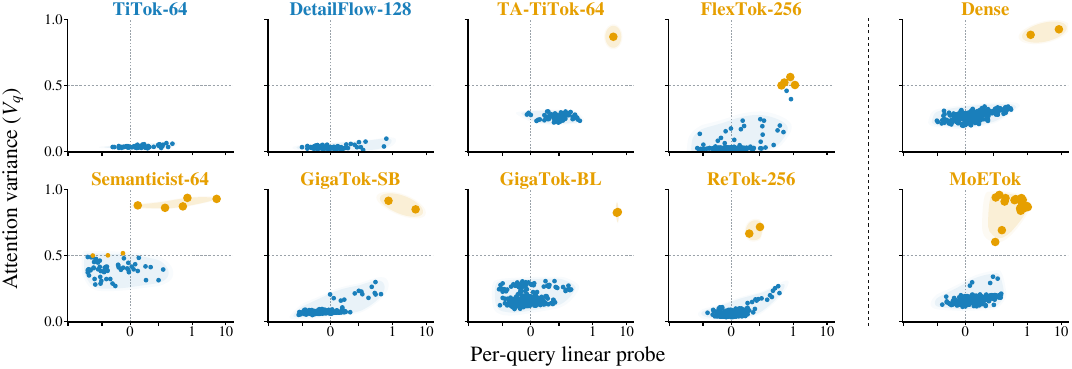}
    \caption{\textbf{Semantic supervision promotes query-role differentiation.} Each query is plotted by attention variance $V_q$ and per-query linear probe score. \detailtext{\textbf{Blue}} and \contenttext{\textbf{orange}} indicate Detail and Content queries; model-name colors distinguish tokenizers without and with semantic supervision.}
    \label{fig:cross-tokenizer-attention-variance}
\end{figure}

\paragraph{Visual representations for understanding and generation.}
Semantic learning improves visual tokenization for reconstruction, representation, and generation~\citep{vavae,maetok,unitok2025,qlip2025,vtp2025,du2025vqrae}. UniTok combines image--text learning and reconstruction with multi-codebook quantization~\citep{unitok2025}; VTP studies joint representation learning and reconstruction for scalable pre-training~\citep{vtp2025}; VQRAE quantizes pretrained visual representations for understanding and generation~\citep{du2025vqrae}. These representations also support unified autoregressive modeling~\citep{wu2024vila,arm2026}. \methodname{} examines the tokenizer's representation space through token roles and conditional computation, linking heterogeneous information to role-specialized processing.

\paragraph{Sparse mixture of experts.}
Sparse mixture-of-experts models expand parameter capacity through input-dependent routing while activating only a subset of experts per token~\citep{moe,switchtransformers}, enabling efficient language-model scaling~\citep{routedscalinglaws,muennighoff2025olmoe}. Vision research has also explored MoE for image classification~\citep{vmoe,softmoe}. Motivated by heterogeneous query roles, we introduce, to our knowledge, the first sparse mixture-of-experts architecture for unified visual tokenizers. It learns role-specialized processing paths for fine-grained visual details and high-level semantic information.

\section{From Heterogeneous Visual Tokens to MoETok}
\label{sec:method}

\subsection{Overview}
\label{sec:method-overview}

Given an image $x$, a 1D visual tokenizer maps it to patch embeddings $P^0=\{p_j^0\}_{j=1}^{N}$ and appends learnable queries $Q^0=\{q_i^0\}_{i=1}^{T}$. A transformer encoder lets the queries gather information from the image patches. Its final query states are projected and quantized into a visual token sequence $z=\{z_i\}_{i=1}^{T}$, which supports image reconstruction and downstream autoregressive modeling. Each query can gather information from the full image, without being assigned to a fixed spatial cell.

\begin{figure*}[t]
    \centering
    \includegraphics[width=\linewidth]{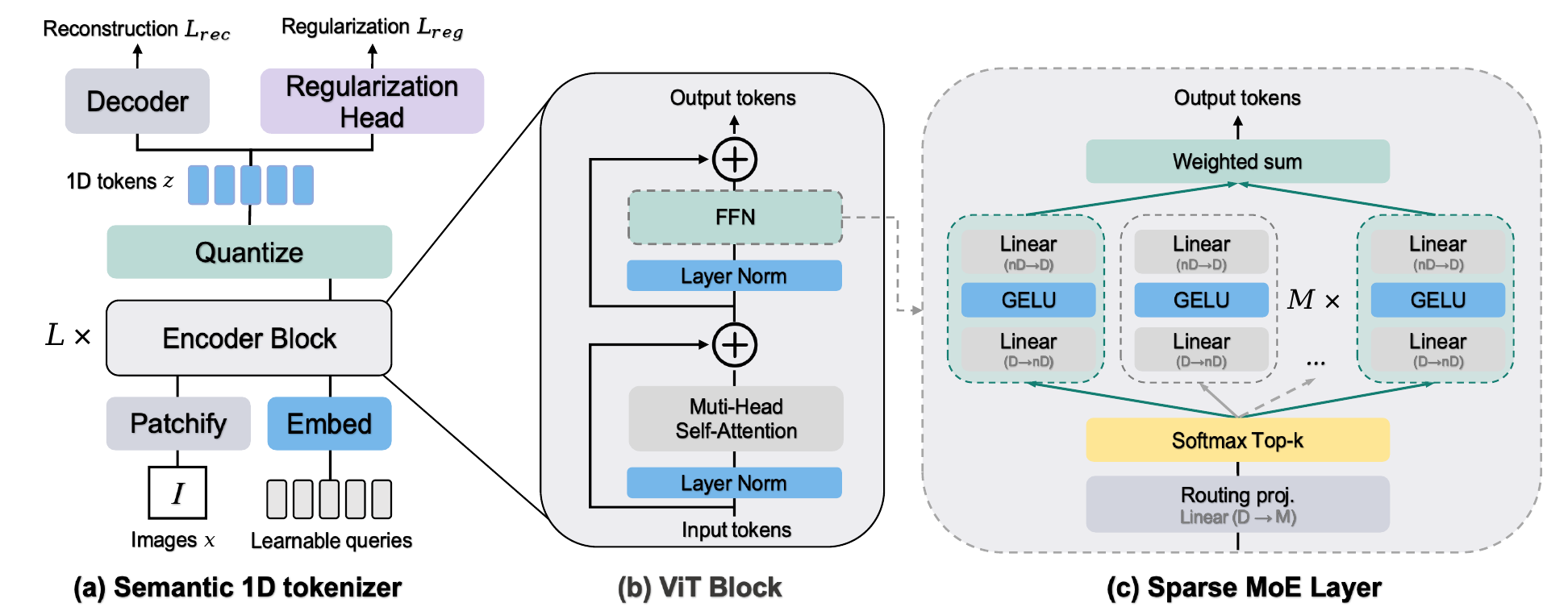}
    \caption{\textbf{MoETok adapts token processing through sparse expert routing.} \textbf{(a)} A 1D tokenizer learns with reconstruction and feature regularization. \textbf{(b)} A standard ViT block. \textbf{(c)} A sparse MoE layer replaces the shared FFN with token-dependent expert selection and weighted aggregation.}
    \label{fig:method-overview}
    \label{fig:moetok-block}
\end{figure*}

\methodname{} introduces token-dependent sparse routing into this architecture. Both patch and query tokens draw from a shared expert pool within each encoder layer (\cref{fig:method-overview}). We first analyze query roles, then introduce the resulting routing design.

\subsection{Emergent Query Roles in Semantic 1D Tokenizers}
\label{sec:token-roles}

Recent 1D tokenizers increasingly incorporate semantic supervision to improve their representations~\citep{gigatok,semanticist,retok2026}. We examine how individual learnable queries gather and encode image information across representative tokenizers, and compare Dense and MoETok under matched reconstruction and feature regularization objectives.

Queries gather image information through attention to patch tokens. Inspired by recent analyses of attention patterns~\citep{doshi2026biorthogonal,marouani2026revisiting}, we examine whether each query attends to stable spatial regions or changes its attention across images. Let $\mathbf A_{\ell,q}(x)$ denote the query-to-patch attention of query $q$ at encoder layer $\ell$ on image $x$, normalized over patch positions, and let $\mathbf U$ denote uniform attention. We define the layer-by-query attention variance as
\begin{equation}
    V_{\ell,q}=
    \frac{
        \operatorname{Var}_{x}\!\left[\mathbf A_{\ell,q}(x)\right]
    }{
        \operatorname{Var}_{x}\!\left[\mathbf A_{\ell,q}(x)\right]
        +\left\|\mathbb E_x\!\left[\mathbf A_{\ell,q}(x)\right]-\mathbf U\right\|_2^2
    }.
    \label{eq:normalized-attention-variance}
\end{equation}
Here, $\operatorname{Var}_x[\mathbf A_{\ell,q}(x)]$ denotes the expected squared deviation from its cross-image mean. The entries form a layer-by-query matrix $\mathbf V=[V_{\ell,q}]\in[0,1]^{L\times T}$. Higher values indicate that cross-image variation dominates the query's stable spatial preference, whereas lower values indicate more consistent attention to similar image regions. For query-level analysis, we average across encoder depth, $V_q=L^{-1}\sum_{\ell=1}^{L}V_{\ell,q}$.

We relate these attention patterns to semantic information through each query's contribution to classification. Following recent work~\citep{beyer2025highly}, we train a linear probe on the complete bottleneck sequence, freeze it, and report the accuracy drop from masking each query as its \emph{per-query linear probe} score.

Without semantic supervision, queries exhibit low attention variance, and most have small per-query classification effects (\cref{fig:cross-tokenizer-attention-variance}). Across models incorporating semantic supervision, a small subset consistently exhibits both high attention variance and strong classification contributions, while most queries remain spatially stable. This recurring pattern indicates that semantic supervision promotes spontaneous query-role differentiation. We use the fixed midpoint $V_q>0.5$ to identify Content queries and refer to the remaining group as Detail queries. We characterize their functional preferences through semantic probing and reconstruction interventions.

\begin{figure}[t]
    \centering
    \newsavebox{\rolefigurethree}
    \newsavebox{\roletablethree}
    \newsavebox{\rolescaledtablethree}
    \sbox{\rolefigurethree}{%
        \includegraphics[width=0.49\linewidth]{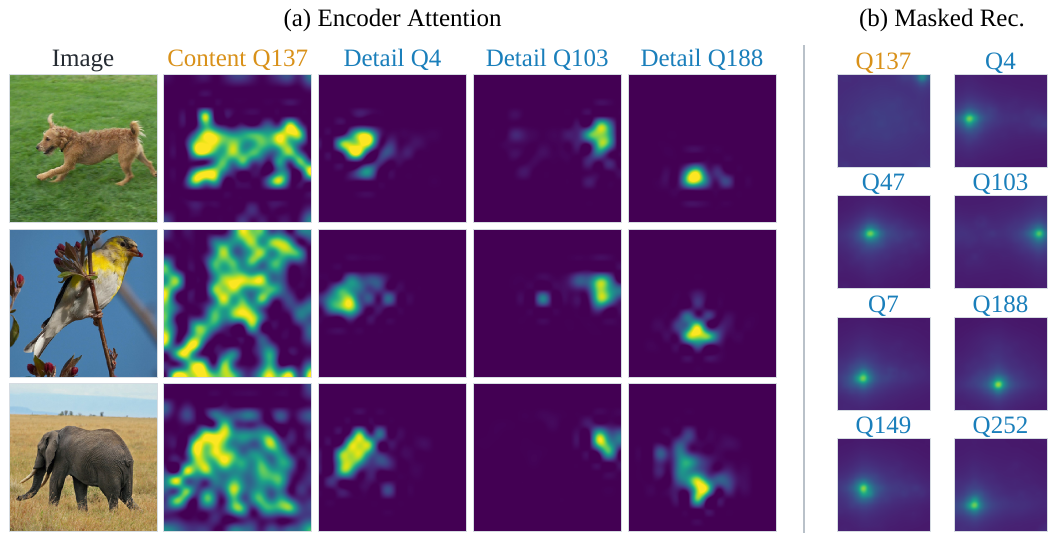}%
    }
    \sbox{\roletablethree}{%
        \scriptsize
        \setlength{\tabcolsep}{1.0pt}
        \renewcommand{\arraystretch}{1.28}
        \newcommand{\metricdeltaactive}[2]{%
            \makebox[6.8em][c]{%
                \textcolor{rolegreentext}{\textbf{#1}}%
                \smash{\makebox[0pt][l]{\ensuremath{{}_{\scriptscriptstyle\color{gray}#2}}}}%
            }%
        }
        \begin{tabular*}{0.485\linewidth}{@{\extracolsep{\fill}}llrcc@{}}
            \toprule
            Model & Subset & \# & Linear probe $\uparrow$ & rFID $\downarrow$ \\
            \midrule
            \multirow{3}{*}{Dense}
              & All & 256 & \cellcolor{rolebase}69.80 & \cellcolor{rolebase}1.52 \\
              & \contenttext{Content} & 2 & \cellcolor{rolegreen}\metricdeltaactive{66.97}{-2.82}
              & \cellcolor{rolered}\textcolor{roleredtext}{\textbf{Failed}} \\
              & \detailtext{Detail} & 254 & 67.66
              & \cellcolor{rolegreen}\metricdeltaactive{1.56}{+0.04} \\
            \midrule
            \multirow{3}{*}{MoETok}
              & All & 256 & \cellcolor{rolebase}74.05 & \cellcolor{rolebase}1.37 \\
              & \contenttext{Content} & 20 & \cellcolor{rolegreen}\metricdeltaactive{76.01}{+1.96}
              & \cellcolor{rolered}\textcolor{roleredtext}{\textbf{Failed}} \\
              & \detailtext{Detail} & 236 & 70.35
              & \cellcolor{rolegreen}\metricdeltaactive{2.08}{+0.71} \\
            \bottomrule
        \end{tabular*}
    }
    \sbox{\rolescaledtablethree}{%
        \resizebox*{!}{0.84589\ht\rolefigurethree}{\usebox{\roletablethree}}%
    }
    \usebox{\rolefigurethree}%
    \hspace{0.025\linewidth}%
    \makebox[0.485\linewidth][c]{%
        \raisebox{\dimexpr 0.01664\ht\rolefigurethree+\dp\rolescaledtablethree\relax}{%
            \usebox{\rolescaledtablethree}%
        }%
    }
    \par\vspace{3pt}
    \begin{minipage}[t]{0.49\linewidth}
        \vspace{0pt}
        \caption{\textbf{Complementary query roles.} \textbf{(a)} \contenttext{\textbf{Content}} attention follows objects; \detailtext{\textbf{Detail}} attention stays localized. \textbf{(b)} Masking Detail queries affects reconstruction in matching regions.}
        \label{fig:query-role-visualization}
    \end{minipage}%
    \hspace{0.025\linewidth}%
    \begin{minipage}[t]{0.485\linewidth}
        \vspace{0pt}
        \captionof{table}{\textbf{Complementary subset functions.} \contenttext{\textbf{Content}} subsets retain strong semantic information; \detailtext{\textbf{Detail}} subsets preserve reconstruction. Content-only decoding fails (rFID~$>300$).}
        \label{tab:query-role-intervention}
    \end{minipage}
    \vspace{-15pt}
\end{figure}
To assess the semantic information retained by each query group, we independently train a linear probe on its mean-pooled final encoder states. Small Content subsets retain strong semantic representations, while Detail subsets carry most of the information needed for reconstruction (\cref{tab:query-role-intervention}). Encoder attention illustrates how Content queries follow foreground objects while Detail queries attend to stable spatial regions (\cref{fig:query-role-visualization}(a)). We then zero-mask each query and average the pixel-wise RGB L2 distance from the full-token reconstruction across 50,000 validation images. Detail queries affect reconstruction in regions matching their attention, whereas the illustrated Content query has little effect (\cref{fig:query-role-visualization}(b)). Together, these interventions validate the connection between attention patterns and complementary semantic and reconstructive functions. Cross-dataset validation and extended visualizations appear in Appendix~\ref{app:attention-variation}.

These findings motivate token-dependent parameter allocation: each token selects a subset of experts from a larger parameter pool, with active computation comparable to dense processing.

\subsection{Sparse Expert Transformer Block}
\label{sec:moetok-block}

Let $X^{\ell}=[P^{\ell};Q^{\ell}]$ denote the concatenated patch and query states entering a routed transformer block. The shared attention sublayer first produces
\begin{equation}
    U^{\ell}=X^{\ell}+\operatorname{Attn}\!\left(\operatorname{LN}(X^{\ell})\right).
\end{equation}
Each layer contains $M$ experts, each implemented as a two-layer FFN with hidden width $nD$, where $D$ is the encoder hidden dimension. For visual token $u_i^{\ell}$, a lightweight router predicts a distribution over experts,
\begin{equation}
    \pi_i^{\ell}=\operatorname{softmax}\!\left(W_r^{\ell}\operatorname{LN}(u_i^{\ell})\right),
    \qquad \mathcal{S}_i^{\ell}=\operatorname{TopK}(\pi_i^{\ell},k),
\end{equation}
where $\mathcal{S}_i^{\ell}\subseteq\{1,\ldots,M\}$ contains the $k$ selected experts. After renormalizing the selected routing weights, the FFN update is
\begin{equation}
    x_i^{\ell+1}=u_i^{\ell}+
    \sum_{e\in\mathcal{S}_i^{\ell}}
    \widetilde{\pi}_{i,e}^{\ell}\,
    E_e^{\ell}\!\left(\operatorname{LN}(u_i^{\ell})\right).
    \label{eq:moetok}
\end{equation}
MoETok replaces the dense FFN in every encoder layer with a sparse MoE layer. Expert assignments are learned from the task objectives, with patch and query tokens sharing a layer-local expert pool. Only the final query states are projected and quantized.

\subsection{Learning Objectives}
\label{sec:objectives}

MoETok is trained with reconstruction, vector-quantization, and visual feature regularization objectives~\citep{gigatok}:
\begin{equation}
    \mathcal{L}=\lambda_{\mathrm{rec}}\mathcal{L}_{\mathrm{rec}}
    +\lambda_{\mathrm{vq}}\mathcal{L}_{\mathrm{vq}}
    +\lambda_{\mathrm{reg}}\mathcal{L}_{\mathrm{reg}}.
    \label{eq:objective}
\end{equation}
We follow GigaTok for the reconstruction objective~\citep{gigatok}. Feature regularization supervises features decoded from the quantized query sequence, with frozen SigLIP2-L features as the default target~\citep{siglip2}. 
We use StableVQ for quantization~\citep{stablevq}.

\subsection{Analyzing Learned Conditional Routing}
\label{sec:routing-analysis}

\begin{figure*}[t]
    \centering
    \includegraphics[width=\textwidth]{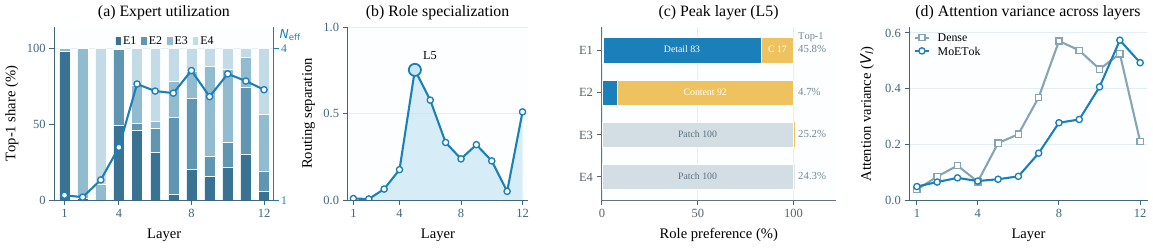}
    \caption{\textbf{Conditional routing learns role-specialized visual processing.} Expert use broadens mid-encoder \textbf{(a)}; role separation peaks mid-encoder and remains high at the output \textbf{(b)}. Experts show distinct role preferences \textbf{(c)}; MoETok retains higher attention variance at the bottleneck \textbf{(d)}.}
    \label{fig:routing-emergence}
\end{figure*}

We analyze Top-1 expert assignments~\citep{muennighoff2025olmoe} to characterize expert utilization and token-role specialization. The entropy-equivalent effective expert count $N_{\mathrm{eff},\ell}\in[1,M]$ ranges from complete concentration to uniform expert use. The normalized Jensen--Shannon divergence between role-conditioned routing distributions measures how distinctly Patch, Detail, and Content tokens select experts. Experts become broadly active from the middle encoder onward, while role specialization peaks at layer 5 and rises again at the output (\cref{fig:routing-emergence}(a, b)). At the peak, distinct experts preferentially process Patch, Detail, and Content tokens (\cref{fig:routing-emergence}(c)), revealing learned role-specialized routing. This depth-dependent utilization persists when the expert pool expands. In 1D tokenizers, loss-free balancing~\citep{deepseekv3} makes expert use more uniform but degrades reconstruction, semantic representation, and generation, supporting task-driven expert allocation across layers. Further comparisons and metric details appear in Appendix~\ref{app:routing-analysis}.

Compared with Dense, MoETok also expands the Content subset while preserving the reconstructive function of Detail queries (\cref{tab:query-role-intervention}). Averaging the attention-variance matrix across queries gives the layer-wise statistic $V_{\ell}=T^{-1}\sum_{q=1}^{T}V_{\ell,q}$ (\cref{fig:routing-emergence}(d)). Attention variance in the Dense model drops sharply at the final encoder layer, whereas MoETok preserves image-adaptive query attention near the discrete bottleneck. Together, these observations link role-specialized routing with image-adaptive attention near the bottleneck, providing an interpretation of the representation gains observed in our controlled comparisons.

\section{Experiments}
\label{sec:experiments}

We first study tokenization structure, performance--compute scaling, and expert design, and then evaluate MoETok on ImageNet and in unified autoregressive modeling.

\subsection{Experimental Setup}
\label{sec:setup}

We train tokenizers on ImageNet~\citep{imagenet} at $256\times256$ resolution. Unless otherwise specified, analyses and controlled studies use 256 queries and four experts with two activated per token. Visual feature regularization uses a frozen SigLIP2-L teacher~\citep{siglip2}. Dense baselines share the same backbone, supervision, and training settings. Tokenizer parameters are reported as active/total (A/T). Detailed architectures and training hyperparameters are provided in Appendix~\ref{app:tokenizer-training}.

We evaluate reconstruction, semantic representation, and downstream generation using rFID, linear probe, zero-shot Top-1 accuracy, and gFID, with a consistent evaluation protocol within each sweep. We train a linear classifier on flattened quantized tokens with the tokenizer frozen. Zero-shot classification compares the tokenizer's aligned visual features with class-name embeddings from the frozen SigLIP2 text encoder. AR models use LlamaGen with 1D RoPE and its B/XL configurations~\citep{llamagen}. Guided ImageNet results use the CFG scale minimizing gFID. Text-conditioned generation is evaluated on GenEval, DPG-Bench, WISE, and GenEval2~\citep{ghosh2023geneval,hu2024ella,niu2026wise,kamath2025geneval2}. Additional protocols and controlled results appear in Appendices~\ref{app:implementation} and~\ref{app:additional-ablation}.

\subsection{Tokenization Structure and Expert Routing}
\label{sec:latent-structure}

\begin{table}[h]
    \centering

    \begin{minipage}{0.82\linewidth}
        \centering
        \caption{\textbf{1D tokenization amplifies the semantic benefits of expert routing.}
        Each pair matches active FFN capacity; gFID is evaluated without guidance.}
        \label{tab:latent-structure}
    \end{minipage}

    \vspace{2pt}

    \resizebox{0.75\linewidth}{!}{
    \begin{tabular}{llrrrr}
        \toprule
        Method & Routed tokens & rFID $\downarrow$ & Linear probe $\uparrow$ & Zero-shot $\uparrow$ & gFID $\downarrow$ \\
        \midrule
        Dense & -- & 1.566 & 41.83 & 61.52 & 12.18 \\
        MoETok & Patch & \textbf{1.494} & \textbf{44.26} & \textbf{63.16} & \textbf{11.59} \\
        \midrule
        Dense & -- & 1.522 & 48.72 & 59.17 & 10.35 \\
        MoETok & Patch + Query & \textbf{1.368} & \textbf{60.22} & \textbf{65.03} & \textbf{9.59} \\
        \bottomrule
    \end{tabular}
    }
\end{table}

We compare sparse routing in patch-based 2D tokenizers and query-based 1D tokenizers, with each dense--MoE pair matched in active FFN capacity, objectives, and training schedule (\cref{tab:latent-structure}). Sparse routing improves reconstruction, linear separability, zero-shot Top-1 accuracy, and generation in both structures. Its benefits are substantially greater for linear separability and zero-shot Top-1 accuracy in the query-based tokenizer, accompanied by a larger reconstruction gain.

Both Dense and MoETok yield lower gFID with query-based than patch-based tokenization, echoing GigaTok's findings on query-based generation~\citep{gigatok}. The larger semantic gains in query-based tokenizers establish 1D tokenization as a particularly effective setting for sparse expert routing, consistent with the heterogeneous query roles identified in our analysis.

\subsection{Performance--Compute Scaling}
\label{sec:scaling}

\noindent
    \begin{minipage}[b]{0.64\linewidth}
        \vspace{0pt}
        \centering
        \includegraphics[width=\linewidth]{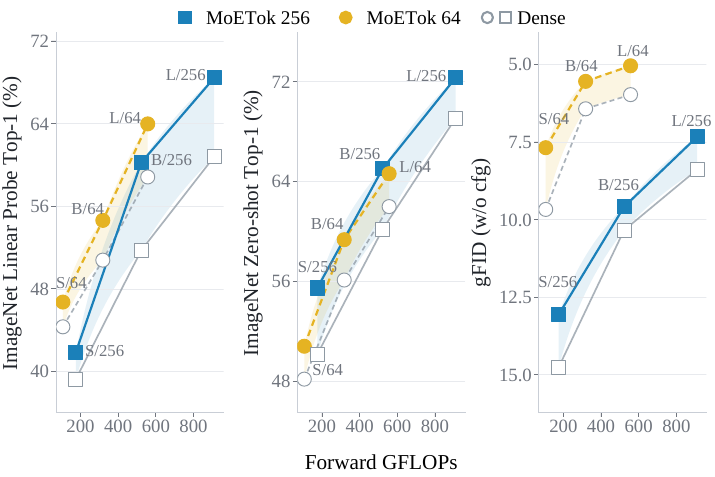}
        \captionof{figure}{\textbf{Performance--compute scaling.} MoETok achieves stronger semantic representation and unguided generation than Dense at comparable forward computation across Small, Base, and Large models at both 64 and 256 tokens.}
        \label{fig:architecture-tokenlength-scaling}
    \end{minipage}
    \hfill
    \begin{minipage}[b]{0.32\linewidth}
        \vspace{0pt}
        Across Small, Base, and Large architectures, MoETok achieves stronger linear probe and zero-shot Top-1 accuracy than Dense at both 64 and 256 tokens (\cref{fig:architecture-tokenlength-scaling}). Our scale variants jointly increase encoder and decoder capacity, while every Dense--MoETok pair shares the same supervision and quantizer. The advantage persists under the fourfold reduction in token count, showing that sparse routing benefits 1D representations at different compression levels. Unguided gFID also improves as model capacity increases, extending the scaling benefits from semantic representation to autoregressive generation.
    \end{minipage}
\par

\subsection{Compact Design Studies}
\label{sec:design-studies}

We examine routing depth and expert configuration in 1D tokenizers.

\noindent
    \begin{minipage}[t]{0.40\linewidth}
        \vspace{0pt}
        Full-depth routing provides the strongest semantic representation with comparable reconstruction quality and is used by default (\cref{tab:layer-coverage}). We vary expert width, expert count, and Top-$k$ to examine per-expert capacity, the available expert pool, and active computation in 1D tokenization (\cref{tab:expert-granularity}).
    \end{minipage}
    \hfill
    \begin{minipage}[t]{0.56\linewidth}
        \vspace{0pt}
        \centering
        \captionof{table}{\textbf{Encoder depth.} Full-depth routing yields the strongest semantic representation while maintaining comparable reconstruction quality.}
        \label{tab:layer-coverage}
        \scriptsize
        \setlength{\tabcolsep}{3.0pt}
        \begin{tabular}{lrrr}
            \toprule
            MoETok layers & rFID $\downarrow$ & Linear probe $\uparrow$ & Zero-shot $\uparrow$ \\
            \midrule
            Early 4 & 1.322 & 46.85 & 59.50 \\
            Late 4 & 1.378 & 51.21 & 61.46 \\
            Late 8 & \textbf{1.309} & 51.31 & 62.30 \\
            All & 1.368 & \textbf{60.22} & \textbf{65.03} \\
            \bottomrule
        \end{tabular}
    \end{minipage}
\par

\noindent\makebox[\linewidth][c]{%
    \begin{minipage}{0.82\linewidth}
    \centering
    \captionof{table}{\textbf{Expert design for 1D tokenizers.} We vary expert width, expert count, and Top-$k$. E8/T2 balances representation, generation, and computational cost.}
    \label{tab:expert-granularity}
    \footnotesize
    \setlength{\tabcolsep}{5pt}
    \renewcommand{\arraystretch}{1.06}
    \resizebox{0.92\linewidth}{!}{%
    \begin{tabular}{lccrrr}
        \toprule
        Experts / Top-$k$ & Width & Total / Active & Linear probe $\uparrow$ & Zero-shot $\uparrow$ & gFID $\downarrow$ \\
        \midrule
        4 / 2 & $2D$ & $8D / 4D$ & 54.01 & 61.82 & 10.47 \\
        4 / 2 & $4D$ & $16D / 8D$ & \textbf{60.22} & \textbf{65.03} & \textbf{9.59} \\
        \arrayrulecolor{gray!40}\midrule\arrayrulecolor{black}
        4 / 2 & $4D$ & $16D / 8D$ & {60.22} & {65.03} & {9.59} \\
        \rowcolor{lightrow}
        8 / 2 & $4D$ & $32D / 8D$ & {60.63} & {66.37} & \textbf{8.99} \\
        16 / 2 & $4D$ & $64D / 8D$ & \textbf{62.92} & \textbf{67.02} & 9.14 \\
        \arrayrulecolor{gray!40}\midrule\arrayrulecolor{black}
        8 / 1 & $4D$ & $32D / 4D$ & 52.02 & 56.65 & 10.56 \\
        \rowcolor{lightrow}
        8 / 2 & $4D$ & $32D / 8D$ & 60.63 & 66.37 & 8.99 \\
        8 / 8 & $4D$ & $32D / 32D$ & \textbf{63.96} & \textbf{67.62} & \textbf{8.66} \\
        \bottomrule
    \end{tabular}}
    \end{minipage}}
\par

Routing the late four layers yields stronger semantic representation than routing the early four, consistent with mid-encoder and output-layer specialization (\cref{fig:routing-emergence}); full-depth routing improves it further. In 1D tokenizers, wider experts improve all three metrics, and expanding the expert pool improves semantic representation at fixed active computation. Eight experts achieve the lowest gFID in this sweep. Increasing Top-$k$ from one to two yields substantial gains; activating all eight provides smaller additional improvements at four times the active expert capacity. Top-2 thus captures most of the gains from full activation. We adopt E8/T2 to balance representation, generation, and computational cost. Further comparisons of routing in the encoder and decoder, along with studies of expert granularity, latent dimension, and codebook size, appear in Appendix~\ref{app:additional-ablation}.

\begin{center}
    \begin{minipage}{\linewidth}
    \centering
    \captionof{table}{\textbf{ImageNet reconstruction, recognition, and generation.} MoETok combines strong semantic representations with high-quality reconstruction and generation in a single 1D tokenizer.}
    \label{tab:imagenet-main}
    \small
    \renewcommand{\arraystretch}{1.02}
    \setlength{\tabcolsep}{3.0pt}
    \resizebox{\linewidth}{!}{
    \begin{tabular}{lcccccccc}
        \toprule
        \multirow[c]{2}{*}{\textbf{Method}} & \multirow[c]{2}{*}{\textbf{Tokens}} & \multirow[c]{2}{*}{\textbf{\makecell{Tok. Params\\A/T}}} & \multicolumn{2}{c}{\textbf{Tokenizer}} & \multicolumn{2}{c}{\textbf{Generator}} & \textbf{w/o CFG} & \textbf{w/ CFG} \\
        \cmidrule(lr){4-5}\cmidrule(lr){6-7}\cmidrule(lr){8-8}\cmidrule(lr){9-9}
        & & & \textbf{rFID} $\downarrow$ & \textbf{Zero-shot} $\uparrow$ & \textbf{Params} & \textbf{Epochs} & \textbf{gFID} $\downarrow$ & \textbf{gFID} $\downarrow$ \\
        \midrule
        \multicolumn{9}{c}{\textit{\textbf{Perception and Generation Baselines}}} \\
        \midrule
        SigLIP2-L~\citep{siglip2} & 256 & 882M & -- & 82.5 & -- & -- & -- & -- \\
        DiT~\citep{dit} & -- & -- & -- & -- & 675M & 1400 & 9.62 & 2.27 \\
        REPA~\citep{repa} & -- & -- & 0.61 & -- & 675M & 800 & 5.78 & 1.29 \\
        DDT~\citep{ddt2025} & -- & -- & 0.61 & -- & 675M & 400 & 6.27 & 1.26 \\
        RAE (SigLIP2-B)~\citep{rae2026} & 256 & 509M & 0.82 & -- & 839M & 80 & 4.93 & -- \\
        FVQ-XL~\citep{fvq2026} & 256 & -- & -- & -- & 775M & -- & -- & 2.07 \\
        \arrayrulecolor{gray!40}\midrule\arrayrulecolor{black}
        VILA-U~\citep{wu2024vila} & $256{\times}4$ & -- & 1.80 & 73.3 & -- & -- & -- & -- \\
        UniTok~\citep{unitok2025} & $256{\times}8$ & -- & 0.38 & 78.4 & 1.4B & -- & 2.51 & 2.77 \\
        QLIP-B~\citep{qlip2025} & 256 & 175M & 3.21 & 74.3 & -- & -- & -- & -- \\
        \midrule
        \multicolumn{9}{c}{\textit{\textbf{1D Visual Tokenizers}}} \\
        \midrule
        TiTok-L~\citep{titok} & 32 & 641M & 2.21 & -- & 177M & -- & 3.15 & 2.77 \\
        Semanticist~\citep{semanticist} & 32 & 768M & 2.63 & -- & 493M & -- & -- & 2.57 \\
        FlexTok d18--d18~\citep{flextok2025} & 32 & 950M & 1.61 & -- & 1.33B & -- & -- & 2.00 \\
        DetailFlow-16~\citep{detailflow2025} & 128 & 271M & 1.22 & -- & 326M & 300 & -- & 2.96 \\
        VFMToK~\citep{vfmtok} & 256 & -- & 0.89 & -- & 343M & 300 & 2.11 & 2.75 \\
        ReTok-S-B~\citep{retok2026} & 256 & 232M & 1.01 & -- & 775M & 300 & -- & 2.27 \\
        GigaTok-B-L~\citep{gigatok} & 256 & 622M & 0.81 & -- & 111M & 300 & -- & 3.26 \\
        GigaTok-XL-XXL~\citep{gigatok} & 256 & 2.9B & 0.79 & -- & 1.4B & 300 & 1.98 & -- \\
        \arrayrulecolor{gray!40}\midrule\arrayrulecolor{black}
        Dense (GigaTok) & 256 & 505M & 0.82 & 78.5 & 111M & 300 & 5.06 & 2.37 \\
        \rowcolor{lightrow}
        & & & & & 111M & 300 & 4.44 & 1.99 \\
        \rowcolor{lightrow}
        \multirow{-2}{*}{\methodname{}} & \multirow{-2}{*}{256} & \multirow{-2}{*}{505M/845M} & \multirow{-2}{*}{\textbf{0.76}} & \multirow{-2}{*}{\textbf{80.4}} & 775M & 300 & \textbf{1.61} & \textbf{1.39} \\
        \bottomrule
    \end{tabular}}
    \end{minipage}
\end{center}

\subsection{ImageNet Reconstruction, Representation, and Generation}
\label{sec:imagenet-main}

Under the same visual feature regularization, MoETok improves reconstruction, semantic representation, and generation over the matched Dense tokenizer (\cref{tab:imagenet-main}). These joint gains demonstrate that sparse expert routing strengthens the learned 1D representation across all three tasks. Stronger semantic representation accompanies improved reconstruction, consistent with the complementary Content and Detail roles identified by our token-level interventions.

Within the 1D tokenizer family, MoETok combines strong reconstruction with state-of-the-art class-conditional generation and the strongest reported zero-shot Top-1 accuracy among the evaluated methods. Its visual tokens support progressively better generation with B and XL LlamaGen priors trained for 300 epochs. With the tokenizer held fixed, this trend complements tokenizer scaling: a larger prior makes more effective use of the same learned visual representation. The following evaluation extends this setting from class-conditional generation to text-conditioned generation and multimodal understanding in a unified autoregressive model.

\subsection{Unified Understanding and Generation}
\label{sec:unified-downstream}

We initialize the unified autoregressive model from Qwen3.5-4B~\citep{qwen35} and extend its vocabulary with MoETok visual tokens. Following discrete multimodal modeling~\citep{wang2024emu3}, we train the model with a next-token cross-entropy objective over text and visual tokens. Training details are provided in Appendix~\ref{app:unified-training}. We evaluate seven standard understanding benchmarks~\citep{li2023evaluating,liu2024mmbench,fu2023mme,yue2024mmmu,hudson2019gqa,goyal2017making,li2024seed} and four generation suites.

The 4B MoETok-based model is competitive with larger discrete models in understanding. It leads evaluated autoregressive baselines on overall GenEval and DPG-Bench and all six WISE domains, and evaluated open models on GenEval2 GM (\cref{tab:unified-understanding,tab:unified-generation,tab:wise,tab:geneval2}). These results demonstrate the versatility of MoETok's 1D representation for unified understanding and generation.

\begin{center}
    \begin{minipage}{\linewidth}
    \centering
    \captionof{table}{\textbf{Multimodal understanding benchmarks.} Unified indicates joint training for understanding and generation; \#Params reports the number of parameters in the language model backbone.}
    \label{tab:unified-understanding}
    \scriptsize
    \setlength{\tabcolsep}{4.0pt}
    \renewcommand{\arraystretch}{0.93}
    \resizebox{\textwidth}{!}{
    \begin{tabular}{lccccccccc}
        \toprule
        Model & Unified & \#Params & POPE & MMB & MME$_{\rm Perc}$ & MMMU & GQA & VQAv2 & SEED \\
        \midrule
        \multicolumn{10}{l}{\emph{Continuous visual representations}} \\
        Qwen3.5-4B~\citep{qwen35} & $\times$ & 4B & -- & 89.4 & -- & 77.6 & -- & -- & -- \\
        Janus-Pro~\citep{chen2025janus} & $\checkmark$ & 7B & 87.4 & 79.2 & 1567 & 41.0 & 62.0 & -- & 72.1 \\
        Show-o2~\citep{xie2025show} & $\checkmark$ & 7B & -- & 79.3 & 1621 & 48.9 & 63.1 & -- & 69.8 \\
        BAGEL~\citep{deng2025emerging} & $\checkmark$ & 7B & -- & 85.0 & 1687 & 55.3 & -- & -- & -- \\
        \midrule
        \multicolumn{10}{l}{\emph{Discrete visual representations}} \\
        Show-o~\citep{xie2024show} & $\checkmark$ & 1.3B & 80.0 & -- & 1097 & 26.7 & 58.0 & 69.4 & -- \\
        Liquid~\citep{wu2026liquid} & $\checkmark$ & 7B & 83.2 & -- & 1448 & -- & 61.1 & 76.8 & -- \\
        VILA-U~\citep{wu2024vila} & $\checkmark$ & 7B & 85.8 & -- & 1402 & -- & 60.8 & 79.4 & 59.0 \\
        UniTok~\citep{unitok2025} & $\checkmark$ & 7B & 83.2 & -- & 1448 & -- & 61.1 & 76.8 & -- \\
        Emu3~\citep{wang2024emu3} & $\checkmark$ & 8B & 85.2 & 58.5 & -- & 31.6 & 60.3 & 75.1 & 68.2 \\
        ARM~\citep{arm2026} & $\checkmark$ & 7B & 87.3 & 80.7 & 1463 & 40.2 & 59.8 & 76.1 & 73.1 \\
        \rowcolor{lightrow}
        \methodname{} & $\checkmark$ & 4B & 85.5 & 75.6 & 1525 & 39.7 & 60.0 & 76.7 & 74.2 \\
        \bottomrule
    \end{tabular}}
    \end{minipage}
\end{center}

\begin{center}
    \begin{minipage}{\linewidth}
    \centering
    \captionof{table}{\textbf{Text-to-image generation on GenEval and DPG-Bench.}}
    \label{tab:unified-generation}
    \scriptsize
    \setlength{\tabcolsep}{4.2pt}
    \renewcommand{\arraystretch}{0.93}
    \resizebox{\textwidth}{!}{
    \begin{tabular}{lccccccc}
        \toprule
        Method & \multicolumn{4}{c}{GenEval $\uparrow$} & \multicolumn{3}{c}{DPG-Bench $\uparrow$} \\
        \cmidrule(lr){2-5}\cmidrule(lr){6-8}
        & Two Obj. & Position & Color Attri. & Overall & Global & Relation & Overall \\
        \midrule
        \multicolumn{8}{l}{\emph{Diffusion-based models}} \\
        \midrule
        FLUX.1-dev~\citep{bfl_flux_2024} & 0.81 & 0.22 & 0.45 & 0.66 & 74.35 & 90.87 & 83.84 \\
        Show-o~\citep{xie2024show} & 0.80 & 0.31 & 0.50 & 0.68 & -- & -- & 67.48 \\
        Lumina-Image 2.0~\citep{qin2025lumina} & 0.87 & -- & 0.62 & 0.73 & -- & 94.85 & 87.20 \\
        BAGEL~\citep{deng2025emerging} & 0.94 & 0.64 & 0.63 & 0.82 & -- & -- & -- \\
        Qwen-Image~\citep{wu2025qwen} & 0.92 & 0.76 & 0.77 & 0.87 & 91.32 & 94.31 & 88.32 \\
        \midrule
        \multicolumn{8}{l}{\emph{Autoregressive models}} \\
        \midrule
        Emu3~\citep{wang2024emu3} & 0.71 & 0.17 & 0.21 & 0.54 & 85.21 & 90.22 & 80.60 \\
        Infinity~\citep{han2025infinity} & 0.85 & 0.49 & 0.57 & 0.73 & 93.11 & 90.76 & 83.46 \\
        Janus-Pro-7B~\citep{chen2025janus} & 0.89 & 0.79 & 0.66 & 0.80 & 86.90 & 89.32 & 84.19 \\
        ARM~\citep{arm2026} & 0.91 & 0.75 & 0.60 & 0.79 & 89.85 & 92.00 & 84.48 \\
        \rowcolor{lightrow}
        \methodname{} & 0.96 & 0.84 & 0.74 & 0.86 & 85.50 & 92.47 & 87.61 \\
        \bottomrule
    \end{tabular}}
    \end{minipage}
\end{center}

\begin{center}
    \centering
    \begin{minipage}[t]{0.62\textwidth}
        \vspace{0pt}
        \centering
        \begin{minipage}[t][2.1\baselineskip][t]{\linewidth}
        \captionof{table}{\textbf{WISE.} World-knowledge-intensive generation across six knowledge domains.}
        \label{tab:wise}
        \end{minipage}
        \fontsize{7.9}{9.0}\selectfont
        \setlength{\tabcolsep}{1.1pt}
        \renewcommand{\arraystretch}{0.93}
        \begin{tabular}{@{}>{\centering\arraybackslash}p{0.065\linewidth}>{\raggedright\arraybackslash}p{0.32\linewidth}*{7}{>{\centering\arraybackslash}p{0.075\linewidth}}@{}}
            \toprule
            Type & Method & Cult. & Time & Space & Bio. & Phys. & Chem. & Avg. \\
            \midrule
            \multirow{4}{*}{Diff.} & Qwen-Image & .63 & .53 & .56 & .34 & .48 & .25 & .51 \\
              & Z-Image & .55 & .47 & .51 & .33 & .48 & .18 & .45 \\
              & BAGEL & .41 & .35 & .31 & .20 & .44 & .26 & .35 \\
              & FLUX.1-dev & .52 & .40 & .53 & .18 & .38 & .24 & .42 \\
            \midrule
            \multirow{3}{*}{AR} & Janus-Pro-7B & .37 & .35 & .28 & .28 & .40 & .23 & .33 \\
              & NextStep-1 & .49 & .38 & .31 & .23 & .47 & .25 & .40 \\
            \rowcolor{lightrow}
              & \methodname{} & .57 & .45 & .47 & .38 & .53 & .38 & .49 \\
            \bottomrule
        \end{tabular}
    \end{minipage}
    \hfill
    \begin{minipage}[t]{0.37\textwidth}
        \vspace{0pt}
        \centering
        \begin{minipage}[t][2.1\baselineskip][t]{\linewidth}
        \captionof{table}{\textbf{GenEval2.} Compositional generation benchmark.}
        \label{tab:geneval2}
        \end{minipage}
        \fontsize{7.9}{9.0}\selectfont
        \setlength{\tabcolsep}{2.0pt}
        \renewcommand{\arraystretch}{0.93}
        \begin{tabular}{@{}>{\raggedright\arraybackslash}p{0.64\linewidth}*{2}{>{\centering\arraybackslash}p{0.13\linewidth}}@{}}
            \toprule
            Method & AM $\uparrow$ & GM $\uparrow$ \\
            \midrule
            Gemini (closed) & 82.80 & 44.60 \\
            \midrule
            SDXL & 50.10 & 9.10 \\
            SD3.5-large & 69.70 & 22.80 \\
            FLUX.1-dev & 67.10 & 21.10 \\
            Qwen-Image & 80.80 & 33.80 \\
            BAGEL + CoT & 70.90 & 23.10 \\
            \rowcolor{lightrow}
            \methodname{} & 77.69 & 40.04 \\
            \bottomrule
        \end{tabular}
    \end{minipage}
\end{center}

\FloatBarrier

\section{Conclusion}
\label{sec:conclusion}

Learnable queries in 1D visual tokenizers develop complementary roles under semantic supervision: Content queries capture high-level semantics, while Detail queries preserve fine-grained visual details. MoETok adapts processing to these roles through sparse expert routing, improving reconstruction, semantic representation, and autoregressive generation across model scales and token budgets. It achieves state-of-the-art class-conditional generation within the 1D tokenizer family and supports competitive understanding and leading generation in unified autoregressive modeling. These results establish token-dependent processing as an effective approach to unified visual representation learning, matching query processing to the needs of reconstruction, understanding, and generation.

\subsection*{AI use statement}

We used AI tools for translation, language polishing, and grammar correction.
The authors reviewed and take responsibility for all content.

\bibliographystyle{iclr2027_conference}
\bibliography{iclr2027_conference}

\clearpage
\appendix
\numberwithin{figure}{section}
\numberwithin{table}{section}
\numberwithin{equation}{section}

\section{Additional Experimental Details}
\label{app:implementation}

\subsection{Tokenizer Architecture and Training}
\label{app:tokenizer-training}

\Cref{tab:app-model-scales,tab:app-tokenizer-training} specify the tokenizer architectures and optimization settings for the main results. For controlled ablations, we train the tokenizer for 20 epochs. We adopt the adversarial training configuration of prior work~\citep{gigatok}. Each controlled dense--MoETok comparison matches the backbone, active FFN capacity, supervision, and training schedule.

\begin{table}[htbp]
    \centering
    \caption{\textbf{Tokenizer model scales.} Model scaling jointly varies encoder and decoder capacity.}
    \label{tab:app-model-scales}
    \small
    \setlength{\tabcolsep}{8pt}
    \begin{tabular}{lccc}
        \toprule
        Configuration & Small & Base & Large \\
        \midrule
        Encoder depth & 8 & 12 & 24 \\
        Encoder heads & 8 & 12 & 16 \\
        Encoder width & 512 & 768 & 1024 \\
        Decoder depth & 12 & 24 & 24 \\
        Decoder heads & 12 & 16 & 16 \\
        Decoder width & 768 & 1024 & 1024 \\
        Latent tokens & 64 / 256 & 64 / 256 & 64 / 256 \\
        MoE layers & \multicolumn{3}{c}{All encoder blocks} \\
        \bottomrule
    \end{tabular}
\end{table}

\begin{table}[htbp]
    \centering
    \caption{\textbf{Tokenizer optimization settings for the main results.}}
    \label{tab:app-tokenizer-training}
    \small
    \setlength{\tabcolsep}{8pt}
    \begin{tabular}{lc}
        \toprule
        Configuration & Setting \\
        \midrule
        Experts / Top-$k$ & 8 / 2 \\
        Latent dimension / codebook size & 256 / 8K \\
        Training epochs & 100 \\
        Global batch size & 512 \\
        Learning-rate schedule & Cosine decay \\
        Base learning rate & $2\times10^{-4}$ \\
        Minimum learning rate & $1\times10^{-5}$ \\
        Warm-up iterations & 3,000 \\
        Optimizer & AdamW~\citep{loshchilov2019decoupled} \\
        Optimizer $(\beta_1,\beta_2)$ & $(0.9,0.95)$ \\
        Reconstruction $\ell_2$ weight & 1.0 \\
        Perceptual loss weight & 1.1 \\
        GAN loss weight & 0.1 \\
        GAN start ratio & 0.3 \\
        SigLIP weights (patch / global) & 1.0 / 0.1 \\
        \bottomrule
    \end{tabular}
\end{table}

\paragraph{Linear probe evaluation.}
For standard tokenizer evaluation, we train a linear classifier~\citep{vtp2025} on flattened quantized tokens with the tokenizer frozen.

\paragraph{Autoregressive evaluation.}
To probe each controlled tokenizer's downstream capabilities, we train an autoregressive model with the LlamaGen GPT-B architecture~\citep{llamagen} for 50,000 iterations, using the same data and optimization settings across tokenizers. We evaluate generation using gFID at CFG scale 1.
For the full ImageNet evaluation, we sweep the CFG scale from 1 to 16 in increments of 0.5 and report the lowest gFID for each model alongside its unguided result.

\FloatBarrier
\subsection{Unified Autoregressive Model Training}
\label{app:unified-training}

We initialize the language model from Qwen3.5-4B~\citep{qwen35}, extend its vocabulary with MoETok visual tokens, and jointly train understanding and generation through next-token cross-entropy~\citep{wang2024emu3}. Pretraining is followed by supervised fine-tuning (SFT). \Cref{tab:app-unified-pretraining,tab:app-unified-sft} report the two stages; batch size is measured in packed text and visual tokens per training iteration.

\begin{table}[htbp]
    \centering
    \caption{\textbf{Unified-model pretraining.} Understanding and generation share one autoregressive model and visual vocabulary.}
    \label{tab:app-unified-pretraining}
    \small
    \renewcommand{\arraystretch}{1.06}
    \begin{tabular}{@{}p{0.47\linewidth}p{0.47\linewidth}@{}}
        \toprule
        Configuration & Pretraining \\
        \midrule
        Language-model initialization & Qwen3.5-4B \\
        Visual tokenizer & MoETok E8/T2; dimension 256; codebook 8K \\
        Visual token count & 256 \\
        Generation / understanding loss weights & $1.5:1.0$ \\
        Generation / understanding token ratio & $60\%:40\%$ \\
        Peak learning rate & $10^{-4}$ \\
        Warm-up iterations & 5,000 \\
        Learning-rate schedule & Constant after warm-up \\
        Optimizer & AdamW~\citep{loshchilov2019decoupled} \\
        Optimizer $(\beta_1,\beta_2,\epsilon)$ & $(0.9,0.95,10^{-15})$ \\
        Weight decay & 0 \\
        Gradient clipping & 1.0 \\
        Input image resolution & $256\times256$ \\
        Iterations & 100,000 \\
        Global batch size & 1,048,576 packed tokens / iteration \\
        Data mixture & Listed below \\
        \bottomrule
    \end{tabular}
\end{table}

\paragraph{Pretraining data.}
Generation data comprise LAION-5B~\citep{schuhmann2022laion5b}, DataComp~\citep{datacomp2024}, JourneyDB~\citep{sun2023journeydb}, TextAtlas5M~\citep{wang2025textatlas5m}, and PosterCraft~\citep{chen2026postercraft}. Understanding data comprise LLaVA-OneVision-1.5-Mid-Training~\citep{an2025llavaonevision15}, BigDocs~\citep{rodriguez2025bigdocs}, Honey-Data-15M~\citep{zhang2026bee}, and MAmmoTH-VL~\citep{guo2025mammothvl}.

\begin{table}[htbp]
    \centering
    \caption{\textbf{Unified-model supervised fine-tuning.} SFT increases the relative weight and token share of understanding tasks.}
    \label{tab:app-unified-sft}
    \small
    \renewcommand{\arraystretch}{1.06}
    \begin{tabular}{@{}p{0.47\linewidth}p{0.47\linewidth}@{}}
        \toprule
        Configuration & Supervised fine-tuning \\
        \midrule
        Language-model initialization & Pretraining checkpoint at 100,000 iterations \\
        Visual tokenizer & MoETok E8/T2; dimension 256; codebook 8K \\
        Visual token count & 256 \\
        Generation / understanding loss weights & $1.0:1.5$ \\
        Generation / understanding token ratio & $40\%:60\%$ \\
        Peak learning rate & $5\times10^{-5}$ \\
        Warm-up iterations & 100 \\
        Learning-rate schedule & Cosine decay to $10^{-6}$ \\
        Optimizer & AdamW~\citep{loshchilov2019decoupled} \\
        Optimizer $(\beta_1,\beta_2,\epsilon)$ & $(0.9,0.95,10^{-15})$ \\
        Weight decay & 0 \\
        Gradient clipping & 1.0 \\
        Input image resolution & $256\times256$ \\
        Iterations & 19,073 \\
        Global batch size & 1,048,576 packed tokens / iteration \\
        Data mixture & Listed below \\
        \bottomrule
    \end{tabular}
\end{table}

\paragraph{SFT data.}
Generation data comprise Fine-T2I~\citep{ma2026finet2i}, JourneyDB~\citep{sun2023journeydb}, BLIP3o-60k~\citep{chen2025blip3}, ShareGPT-4o-Image~\citep{chen2025sharegpt4oimage}, Echo-4o-Image~\citep{ye2025echo4o}, TextAtlas5M~\citep{wang2025textatlas5m}, and PosterCraft~\citep{chen2026postercraft}. Understanding data comprise LLaVA-OneVision-1.5-Instruct~\citep{an2025llavaonevision15} and MAmmoTH-VL~\citep{guo2025mammothvl}.

\paragraph{WISE evaluation.}
All results in \cref{tab:wise} use the WISE\_Verified protocol with the Qwen3.5-35B-A3B judge~\citep{niu2026wise}.

\FloatBarrier
\section{Token-Role and Routing Analyses}
\label{app:token-role-analysis}

\subsection{Attention Variance}
\label{app:query-role-identification}
\label{app:attention-variation}

We extend the main-text analysis across image distributions, encoder layers, and the complete query sequence. Attention maps are averaged across heads and normalized over patch positions before computing $V_{\ell,q}$ as defined in Eq.~\eqref{eq:normalized-attention-variance}. For the degenerate case of uniform attention on every image, we set $V_{\ell,q}=0$. DetailFlow-128~\citep{detailflow2025} in \cref{fig:cross-tokenizer-attention-variance} is trained without global SigLIP supervision.

\begin{figure}[!htbp]
    \centering
    \includegraphics[width=\linewidth]{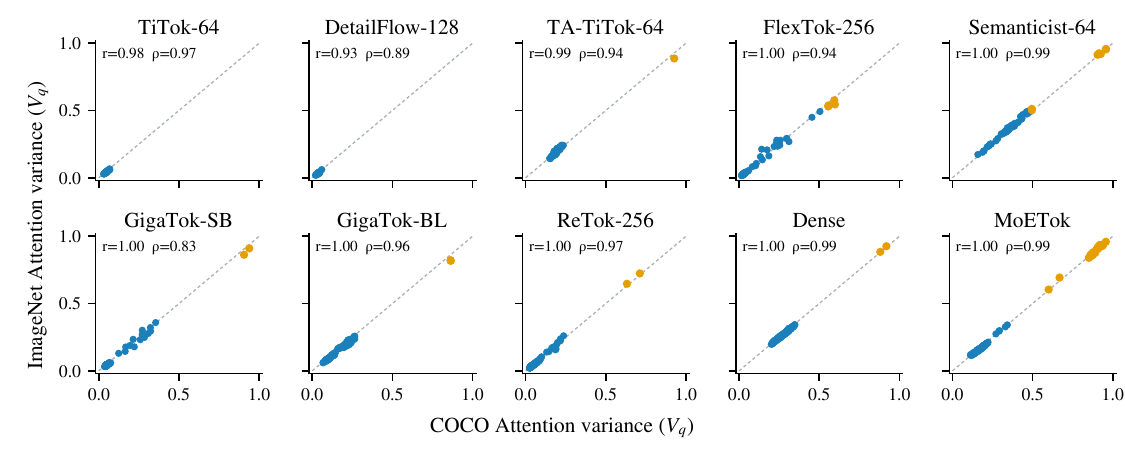}
    \caption{\textbf{Query-level attention variance is consistent across datasets.} Each point compares the same query on COCO and ImageNet; $r$ and $\rho$ denote Pearson and Spearman correlations.}
    \label{fig:app-cross-dataset-attention-variance}
\end{figure}

\paragraph{Cross-dataset stability.}
We compute query-level attention variance $V_q$ independently on 50,000 ImageNet validation images and 30,000 COCO images~\citep{lin2014coco}. Across the ten tokenizers in \cref{fig:app-cross-dataset-attention-variance}, Pearson correlations range from 0.93 to 1.00 and Spearman correlations from 0.83 to 0.99. The strong linear and rank correlations show that both the values and relative ordering of query attention variance remain consistent across these image distributions.

\begin{figure}[!htbp]
    \centering
    \includegraphics[width=\textwidth]{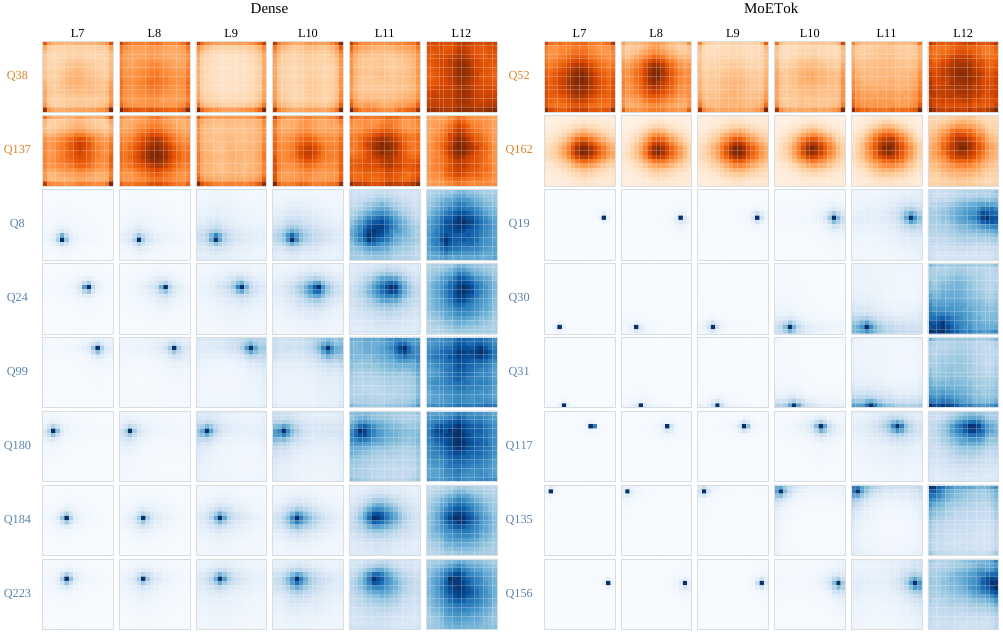}
    \caption{\textbf{Content and Detail attention patterns across encoder depth.} MoETok preserves clearer role differentiation near the bottleneck than Dense.}
    \label{fig:app-query-attention-atlas}
\end{figure}

\paragraph{Attention patterns across depth.}
\Cref{fig:app-query-attention-atlas} follows selected queries through the final six encoder layers. The first two rows for each model show Content queries, and the remaining rows show Detail queries. Both models are evaluated on the same images; each head-averaged map is peak-normalized to highlight its spatial pattern. The layer-wise comparison reveals how the two roles evolve toward the bottleneck: their attention patterns become less distinct at the Dense output, while MoETok maintains clearer differentiation.

\begin{figure}[!htbp]
    \centering
    \includegraphics[width=\textwidth]{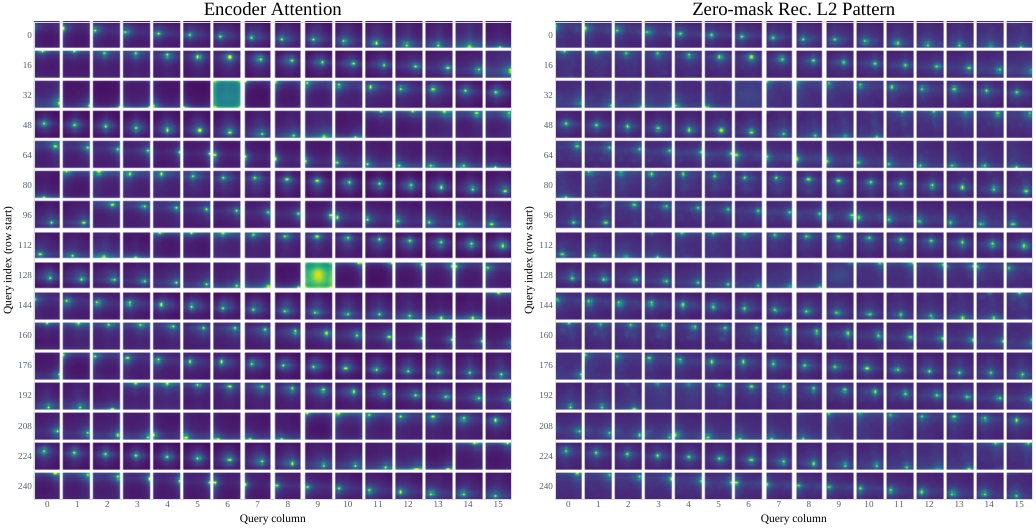}
    \caption{\textbf{Encoder attention aligns with reconstruction contributions.} Detail queries exhibit highly consistent spatial patterns in encoder attention (left) and reconstruction changes under zero masking (right). Content queries with image-adaptive attention have negligible reconstruction effects.}
    \label{fig:app-dense-attention-maskrec}
\end{figure}

\paragraph{Attention and reconstruction across queries.}
\Cref{fig:app-dense-attention-maskrec} extends the single-query masking analysis~\citep{beyer2025highly} to all 256 Dense queries. Matched grid cells show each query's encoder attention and the pixel-wise RGB L2 change induced by zero-masking its quantized token. Detail queries have strongly aligned attention and reconstruction footprints, whereas Content queries attend adaptively across images and induce negligible reconstruction changes. This full-sequence view establishes that the correspondence observed in the main-text examples extends across the query set.

\subsection{Layer-wise Routing Analysis}
\label{app:routing-analysis}

Following category-conditioned routing analyses~\citep{muennighoff2025olmoe}, we measure $p_\ell(e\mid c)$, the fraction of tokens of role $c$ assigned to expert $e$ by Top-1 routing. The overall assignment distribution $p_\ell(e)$ pools all tokens with their observed frequencies. Panel (a) of \cref{fig:routing-emergence} reports its entropy-equivalent effective expert count, $N_{\mathrm{eff},\ell}=\exp H(p_\ell)$~\citep{hill1973diversity}, where $H$ is Shannon entropy. Panel (b) reports the generalized Jensen--Shannon divergence among the three role-conditioned distributions, using equal role weights and dividing by $\log 3$~\citep{lin1991divergence}. All logarithms are natural; the normalized score lies in $[0,1]$, with larger values indicating stronger routing separation.

Panel (c) normalizes $p_\ell(e\mid c)$ across roles within each expert to show role-balanced preferences. The percentages at right give overall Top-1 assignment shares; expert indices are layer-local.

\begin{figure}[!htbp]
    \centering
    \includegraphics[width=0.95\linewidth]{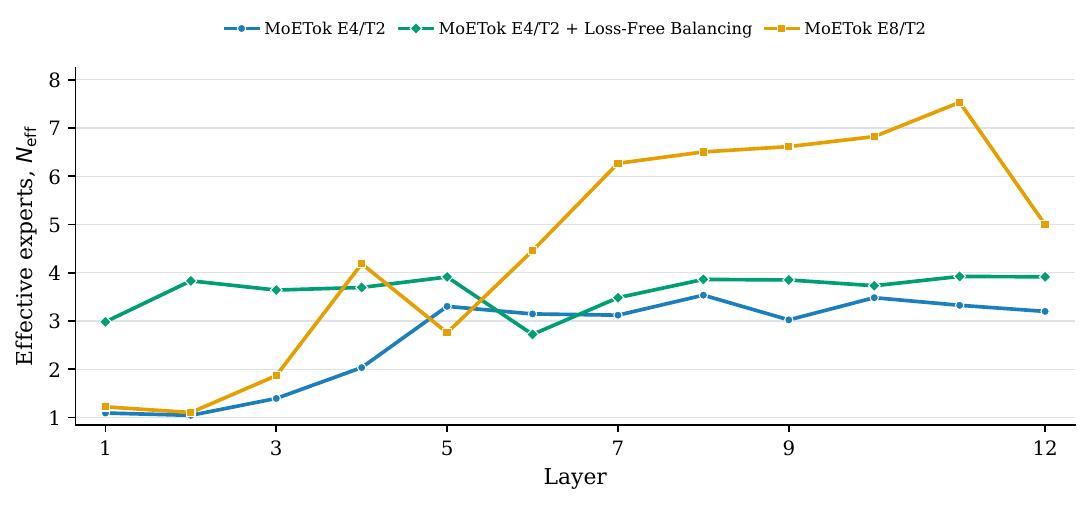}
    \caption{\textbf{Expert utilization across depth in 1D tokenizers.} Both expert-pool sizes exhibit concentrated early-layer routing and broader utilization in deeper layers. Loss-free balancing makes utilization more uniform across most layers.}
    \label{fig:app-expert-utilization}
\end{figure}

\paragraph{Depth-dependent expert utilization.}
Both E4/T2 and E8/T2 concentrate expert use in early layers and engage a broader expert pool in middle and late layers (\cref{fig:app-expert-utilization}). We additionally evaluate loss-free balancing~\citep{deepseekv3} with E4/T2. Balancing moves $N_{\mathrm{eff}}$ closer to its maximum of four across most layers, but worsens reconstruction, semantic representation, and generation (\cref{tab:app-loss-free-balancing}). These results support allowing tokenizer objectives to determine expert allocation across depth, rather than encouraging uniform utilization in every layer.

\begin{table}[!htbp]
    \centering
    \caption{\textbf{Loss-free balancing in a 1D tokenizer.} For E4/T2, task-driven routing achieves better reconstruction, semantic representation, and generation than the balancing variant.}
    \label{tab:app-loss-free-balancing}
    \small
    \setlength{\tabcolsep}{6pt}
    \begin{tabular}{lrrrr}
        \toprule
        Loss-free balancing & rFID $\downarrow$ & Linear probe $\uparrow$ & Zero-shot $\uparrow$ & gFID $\downarrow$ \\
        \midrule
        \rowcolor{lightrow}
        Disabled & \textbf{1.368} & \textbf{60.22} & \textbf{65.03} & \textbf{9.59} \\
        Enabled & 1.491 & 55.39 & 63.29 & 12.65 \\
        \bottomrule
    \end{tabular}
\end{table}

\FloatBarrier
\section{Additional Controlled Studies}
\label{app:additional-ablation}

\paragraph{Encoder and decoder routing.}
In 1D tokenizers, routing in the encoder substantially improves semantic representation and achieves the best reconstruction and generation among the four configurations (\cref{tab:app-encoder-decoder-routing}). Routing in the decoder alone provides modest semantic gains but worsens reconstruction. Routing in both further improves semantic representation, while yielding higher rFID and gFID than routing in the encoder alone. These results support our design of applying expert routing throughout the encoder while retaining a dense decoder.

\begin{table}[!htbp]
    \centering
    \caption{\textbf{Encoder and decoder routing in 1D tokenizers.} Routing in the encoder gives the best reconstruction and generation; extending routing to the decoder further improves semantic representation.}
    \label{tab:app-encoder-decoder-routing}
    \small
    \setlength{\tabcolsep}{5pt}
    \renewcommand{\arraystretch}{1.02}
    \begin{tabular}{llrrrr}
        \toprule
        Encoder & Decoder & rFID $\downarrow$ & Linear probe $\uparrow$ & Zero-shot $\uparrow$ & gFID $\downarrow$ \\
        \midrule
        Dense & Dense & 1.522 & 48.72 & 59.17 & 10.35 \\
        \rowcolor{lightrow}
        MoE & Dense & \textbf{1.368} & 60.22 & 65.03 & \textbf{9.59} \\
        Dense & MoE & 1.813 & 49.41 & 60.39 & 9.95 \\
        MoE & MoE & 1.458 & \textbf{62.75} & \textbf{66.61} & 11.23 \\
        \bottomrule
    \end{tabular}
\end{table}

\paragraph{Routing granularity in 1D tokenizers.}
We jointly vary expert count, width, and Top-$k$ while keeping total and active expert capacities fixed (\cref{tab:app-routing-granularity}). E4/T2 outperforms both fewer, wider experts and more numerous, narrower experts across all three metrics. The allocation of a fixed expert capacity budget thus affects representation and generation in 1D tokenizers.

\begin{table}[!htbp]
    \centering
    \caption{\textbf{Routing granularity in 1D tokenizers.} Total and active expert capacities are fixed at $16D$ and $8D$. E4/T2 performs best across all three metrics.}
    \label{tab:app-routing-granularity}
    \small
    \setlength{\tabcolsep}{5pt}
    \begin{tabular}{lccrrr}
        \toprule
        Experts / Top-$k$ & Width & Total / Active & Linear probe $\uparrow$ & Zero-shot $\uparrow$ & gFID $\downarrow$ \\
        \midrule
        2 / 1 & $8D$ & $16D / 8D$ & 55.08 & 61.69 & 10.80 \\
        4 / 2 & $4D$ & $16D / 8D$ & \textbf{60.22} & \textbf{65.03} & \textbf{9.59} \\
        8 / 4 & $2D$ & $16D / 8D$ & 57.21 & 64.02 & 9.85 \\
        \bottomrule
    \end{tabular}
\end{table}

\paragraph{Latent dimension and codebook size.}
\Cref{tab:bottleneck-design} separates the effects of codebook size and latent dimension. At dimension 256, increasing the codebook from 8K to 16K improves all four metrics. Expanding it further to 64K slightly improves rFID while reducing semantic and generation performance. With an 8K codebook, dimension 256 gives the strongest linear probe and lowest gFID, whereas dimension 1024 favors zero-shot accuracy and reconstruction. These results show that codebook size and latent width affect representation, reconstruction, and generation differently, with intermediate settings offering a favorable overall balance.

\begin{table}[!htbp]
    \centering
    \caption{\textbf{Latent dimension and codebook size.} We vary codebook size at dimension 256 (top) and dimension with an 8K codebook (bottom). gFID uses CFG scale 1.}
    \label{tab:bottleneck-design}
    \small
    \setlength{\tabcolsep}{5pt}
    \renewcommand{\arraystretch}{1.02}
    \begin{tabular}{ccrrrr}
        \toprule
        Dim. & Codebook & Linear probe $\uparrow$ & Zero-shot $\uparrow$ & rFID $\downarrow$ & gFID $\downarrow$ \\
        \midrule
        256 & 8K & 60.63 & 66.37 & 1.53 & 8.99 \\
        256 & 16K & \textbf{62.77} & \textbf{67.48} & 1.40 & \textbf{8.74} \\
        256 & 64K & 62.01 & 67.28 & \textbf{1.39} & 10.32 \\
        \arrayrulecolor{gray!40}\midrule\arrayrulecolor{black}
        64 & 8K & 58.84 & 64.61 & 1.56 & 9.70 \\
        256 & 8K & \textbf{60.63} & 66.37 & 1.53 & \textbf{8.99} \\
        1024 & 8K & 59.38 & \textbf{66.69} & \textbf{1.49} & 9.42 \\
        \bottomrule
    \end{tabular}
\end{table}

\FloatBarrier

\end{document}